\documentclass[10pt,twocolumn]{article}

\usepackage[utf8]{inputenc}
\usepackage[T1]{fontenc}
\usepackage{amsmath,amssymb,amsfonts}
\usepackage{graphicx}
\graphicspath{{figures/}}
\usepackage{booktabs}
\usepackage{multirow}
\usepackage{hyperref}
\usepackage[margin=1in]{geometry}
\usepackage{caption}
\usepackage{subcaption}
\usepackage{xcolor}
\usepackage{colortbl}
\usepackage{array}
\usepackage{tabularx}
\usepackage{algorithm}
\usepackage{algpseudocode}
\usepackage{enumitem}
\usepackage{float}
\usepackage{pifont}

\definecolor{routeblue}{RGB}{52, 119, 180}
\definecolor{gaingreen}{RGB}{39, 139, 64}
\definecolor{hurtred}{RGB}{192, 57, 43}
\definecolor{neutralgray}{RGB}{127, 127, 127}

\title{\textbf{Can Small Language Models Know What They Don't Know?}\\
\large Semantic Entropy as a Confidence Signal for Sub-3B Parameter Models}

\author{
Prashant Mudgal \\
\texttt{prashant.mudgal@nagarro.com}\\
\texttt{prashantmdgl9@gmail.com}
}

\date{}

\begin{document}
\maketitle

\begin{abstract}
We explore whether entropy-based confidence signals can be leveraged to improve the accuracy of Small Language Models (SLMs) with fewer than 3 billion parameters, running entirely on consumer hardware. We evaluate seven distinct approaches, including token-level entropy early stopping, semantic entropy estimation, and uncertainty-aware routing to larger expert models, across 7 model pairs and 5 standard NLU benchmarks. Our key finding is that \textbf{token-level entropy is effectively blind in SLMs}: in 91\% of dataset--model combinations, mean token entropy is near zero regardless of answer correctness, rendering token-based confidence signals unusable at this scale. We demonstrate that \textbf{semantic entropy}, computed by generating multiple samples, clustering answers by meaning, and measuring distributional uncertainty, recovers a viable confidence signal. Using semantic entropy to selectively route uncertain queries to a larger expert model yields accuracy improvements of up to +50 percentage points. Notably, cross-family routing (e.g., SmolLM~360M to Phi-3.5-mini) averages +22.0\% improvement compared to +6.8\% for same-family routing, revealing that expert model quality matters more than architectural compatibility. Our results suggest that the value proposition for entropy-based methods in SLMs is not computational savings but \emph{intelligent compute allocation}: spending more tokens where they matter most.
\end{abstract}

\section{Introduction}

Large Language Models (LLMs) with tens or hundreds of billions of parameters have demonstrated remarkable reasoning capabilities, and Shannon entropy computed from token-level log-probabilities has been explored as a confidence signal for guiding inference decisions~\cite{sharma2025tjue, han2025overthinking}. However, the practical deployment landscape increasingly demands capable models that can run on consumer-grade hardware such as laptops, edge devices, and single-GPU workstations, where models are constrained to fewer than 7 to 8 billion parameters.

This raises a natural question: \textbf{do entropy-based confidence signals transfer to Small Language Models?}

We set out to answer this question through an exploratory study. Rather than proposing a single method and evaluating it, we systematically compare seven different confidence-based approaches across diverse model families and scales, seeking to understand what works, what fails, and why.

Our investigation yields three principal findings:

\begin{enumerate}[leftmargin=*]
    \item \textbf{Token-level entropy is blind in SLMs.} In 32 out of 35 dataset--model combinations (91\%), the mean token entropy is effectively zero ($<0.01$ bits). SLMs produce short, maximally confident outputs regardless of correctness. Token-level confidence signals do not emerge at this scale.

    \item \textbf{Semantic entropy recovers a useful uncertainty signal.} By generating multiple samples and measuring disagreement at the \emph{answer level} rather than the \emph{token level}, we can identify when an SLM is uncertain. This signal enables selective routing to a larger expert model, improving accuracy by up to +50 percentage points.

    \item \textbf{Expert quality matters more than architectural compatibility.} Cross-family routing to a high-quality expert (e.g., SmolLM~360M to Phi-3.5-mini) yields substantially larger gains (+26.0\% average) than same-family routing to a weaker sibling model (+3.0\% average). The routing mechanism itself is not the bottleneck; the expert's absolute capability is.
\end{enumerate}

Importantly, these accuracy gains come at the cost of \emph{more} tokens, not fewer: semantic approaches require generating multiple samples, and routing invokes a second, larger model. In all 35 dataset--model combinations, semantic routing used more tokens than the baseline. The value proposition shifts from ``spend less'' to ``spend smarter,'' allocating additional compute precisely where the small model is uncertain.

\begin{figure*}[t]
    \centering
    \includegraphics[width=0.95\textwidth]{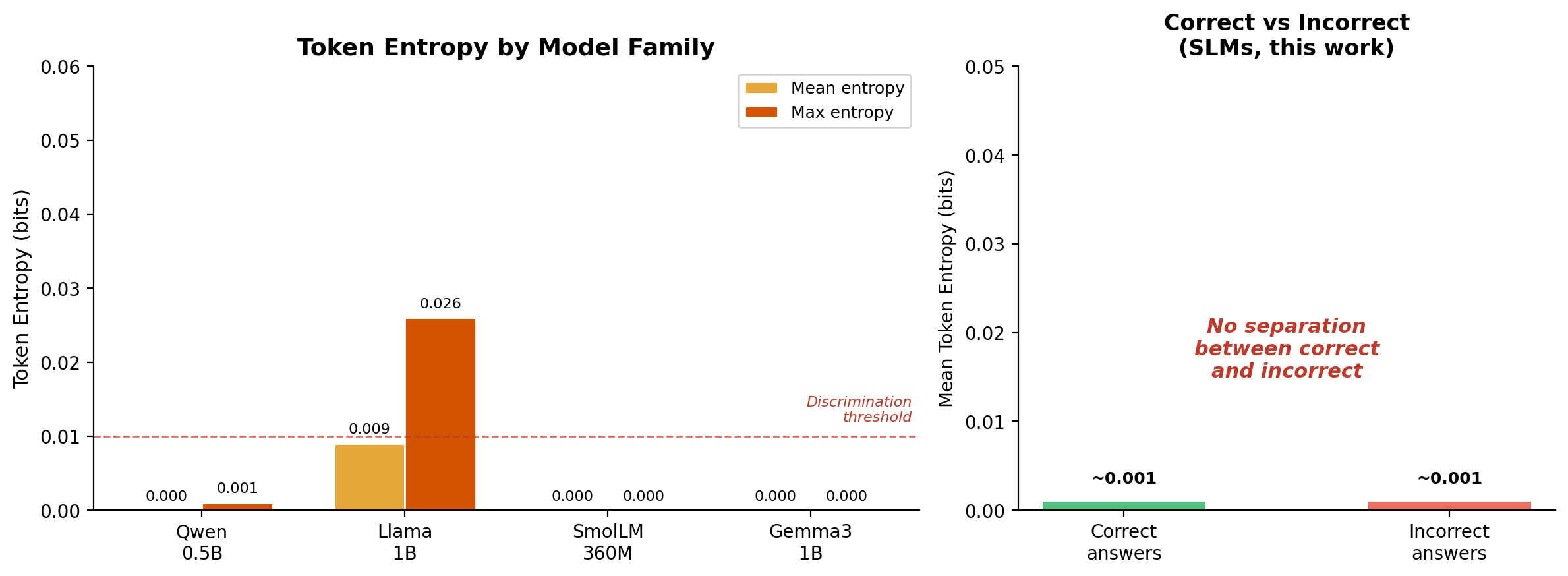}
    \caption{\textbf{Token-level entropy is blind in SLMs.} \textbf{Left:} Mean and maximum token entropy aggregated by model family across all 5 datasets. Three of four families produce entropy below 0.001 bits; only Llama-1B shows any measurable variation (max 0.026 bits), still well below a useful discrimination threshold. \textbf{Right:} Correct and incorrect answers produce identical near-zero mean token entropy, making threshold-based confidence discrimination impossible at this scale.}
    \label{fig:token_entropy_blind}
\end{figure*}

\section{Background}

\subsection{Shannon Entropy as a Confidence Signal}

Shannon entropy~\cite{shannon1948} quantifies the uncertainty in a probability distribution. For a language model generating token $t$ with a distribution over the vocabulary, the entropy is:
\begin{equation}
    H = -\sum_{i=1}^{k} p_i \log_2 p_i
\end{equation}
where $p_i$ are the probabilities of the top-$k$ tokens after softmax normalization. A lower entropy indicates higher confidence (the model concentrates probability mass on fewer tokens), while higher entropy indicates uncertainty (probability is spread across many tokens).

For a generated sequence of $T$ tokens, the mean entropy provides an aggregate confidence signal:
\begin{equation}
    H_{\text{mean}} = \frac{1}{T} \sum_{t=1}^{T} H_t
\end{equation}

This signal has been explored for guiding inference in large reasoning models~\cite{sharma2025tjue, han2025overthinking}, but its behavior in small models ($<$4B parameters) remains largely unexplored.

\subsection{Semantic Entropy}

Semantic entropy~\cite{kuhn2023semantic} addresses a fundamental limitation of token-level entropy: a model may be confident about each individual token it produces while being uncertain about the overall \emph{meaning} of its answer. Semantic entropy operates at the answer level:

\begin{enumerate}[leftmargin=*]
    \item Generate $N$ independent samples from the model (using temperature sampling)
    \item Cluster the answers by semantic meaning
    \item Compute Shannon entropy over the cluster probability distribution
\end{enumerate}

For $C$ meaning clusters with probabilities $\{q_1, q_2, \ldots, q_C\}$:
\begin{equation}
    H_{\text{semantic}} = -\sum_{c=1}^{C} q_c \log_2 q_c
\end{equation}

When all samples agree (1 cluster), $H_{\text{semantic}} = 0$. When samples are maximally scattered ($N$ clusters for $N$ samples), entropy reaches $\log_2 N$. For $N=5$ samples, the semantic entropy ranges from 0.0 to 2.32 bits.

\subsection{Model Cascading}

Model cascading (or routing) is a strategy where queries are first processed by a smaller, cheaper model, and selectively forwarded to a larger, more capable model when the smaller model's confidence is low~\cite{chen2023frugalgpt}. Our work uses entropy-based confidence as the routing signal.

\section{Methodology}

\subsection{Approaches Evaluated}

\begin{figure*}[t]
    \centering
    \includegraphics[width=0.92\textwidth]{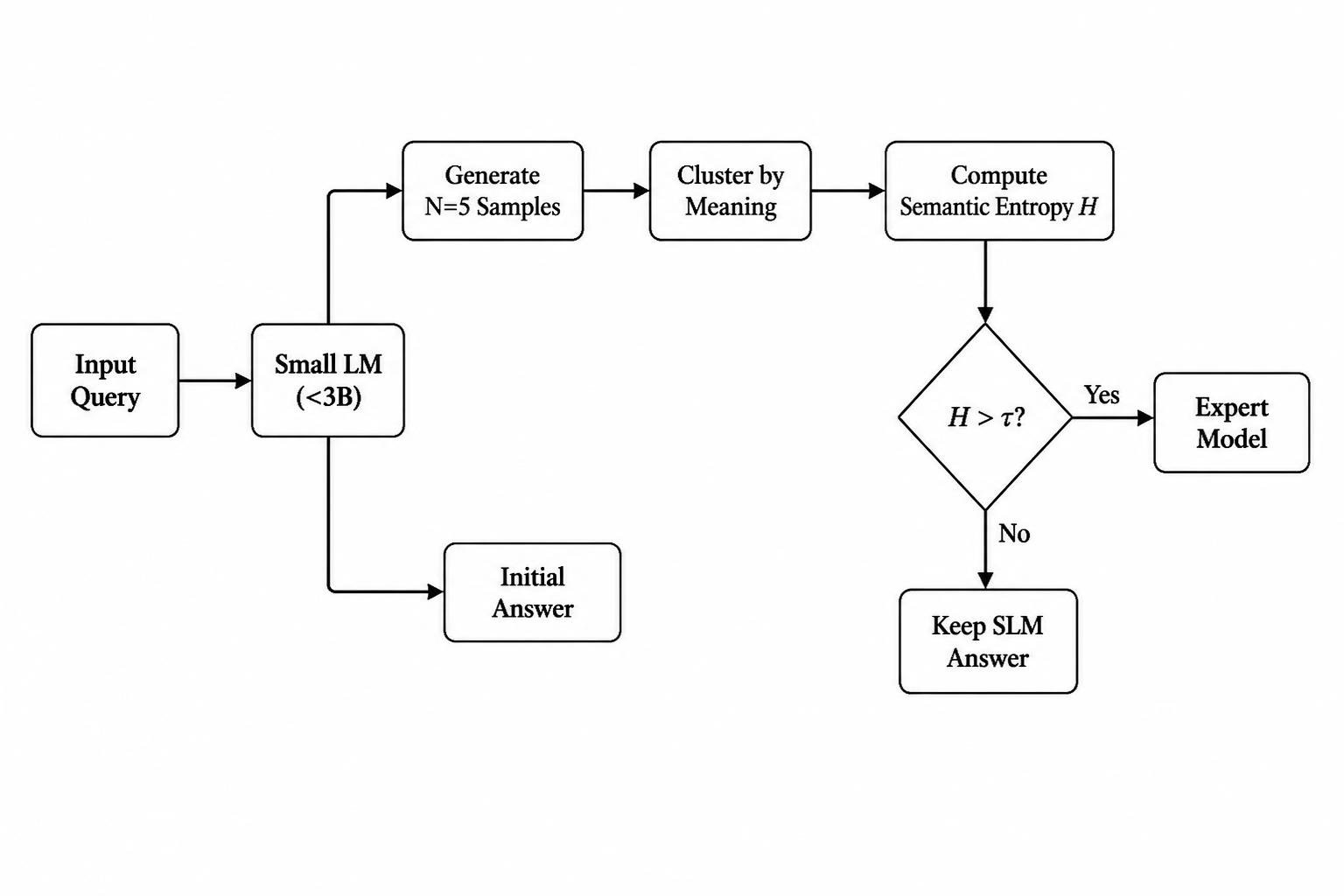}
    \caption{\textbf{Semantic entropy routing framework.} Given an input query, the SLM generates $N=5$ independent samples via temperature sampling. Answers are clustered by meaning (exact match for MCQ; agglomerative clustering for free-text). Shannon entropy is computed over the cluster distribution. If semantic entropy exceeds the calibrated threshold~$\tau$ (set to the mean entropy of correctly answered calibration examples), the query is routed to a larger expert model. Otherwise, the SLM's dominant-cluster answer is accepted.}
    \label{fig:framework}
\end{figure*}

We evaluate seven approaches that combine different confidence signals with different actions:

\begin{enumerate}[leftmargin=*]
    \item \textbf{Baseline}: Generate a single response from the small model with no optimization. This represents the default inference behavior.

    \item \textbf{Token Entropy Early Stop}: Compute the mean Shannon entropy across generated tokens. If entropy falls below a calibrated threshold $\tau$, accept the short response. Otherwise, allow the model to continue generating up to the maximum token limit.

    \item \textbf{MaxProb Early Stop}: Use the mean maximum token probability (highest softmax probability per position) as the confidence signal instead of entropy.

    \item \textbf{ProbGap Early Stop}: Use the mean probability gap ($p_1 - p_2$, difference between the top-two token probabilities) as the confidence signal.

    \item \textbf{Token Entropy Routing}: Compute token-level entropy. If entropy exceeds the threshold (indicating low confidence), route the query to a larger expert model instead of using the small model's answer.

    \item \textbf{Semantic Entropy Early Stop}: Generate $N=5$ samples, compute semantic entropy over answer clusters. If semantic entropy is below the threshold, accept the dominant answer. Otherwise, continue with the full generation.

    \item \textbf{Semantic Entropy Routing}: Generate $N=5$ samples and compute semantic entropy. If semantic entropy exceeds the threshold, route the query to a larger expert model. This combines semantic uncertainty detection with model cascading (Figure~\ref{fig:framework}).
\end{enumerate}

\subsection{Threshold Calibration}

For each approach, we calibrate the confidence threshold $\tau$ using a held-out calibration set of 15 examples per dataset. The threshold is set to the mean entropy (or confidence metric) of examples where the small model produces correct answers:
\begin{equation}
    \tau = \frac{1}{|\mathcal{C}|} \sum_{x \in \mathcal{C}} H(x)
\end{equation}
where $\mathcal{C}$ is the set of correctly answered calibration examples. This ensures that the threshold captures the typical confidence level when the model ``knows'' the answer.

\subsection{Answer Clustering for Semantic Entropy}

For multiple-choice questions (MCQ), we extract the answer letter (A/B/C/D) and cluster by exact match. For boolean questions, we extract yes/no. For free-text answers, we use bag-of-tokens embeddings with agglomerative clustering (cosine similarity, average linkage, threshold 0.85).

\subsection{Framework Architecture}

\begin{algorithm}[t]
\caption{Semantic Entropy Routing}
\label{alg:routing}
\begin{algorithmic}[1]
\Require Query $q$, Small model $M_s$, Expert model $M_e$, Threshold $\tau$, Number of samples $N$
\State \textbf{Generate} $N$ samples: $\{a_1, \ldots, a_N\} \leftarrow M_s(q)$
\State \textbf{Extract} core answers from each sample
\State \textbf{Cluster} answers by meaning $\rightarrow C$ clusters
\State \textbf{Compute} $H_{\text{semantic}} = -\sum_{c=1}^{C} q_c \log_2 q_c$
\If{$H_{\text{semantic}} > \tau$}
    \State \textbf{Route}: $\text{answer} \leftarrow M_e(q)$ \Comment{Use expert}
\Else
    \State \textbf{Accept}: $\text{answer} \leftarrow \text{dominant cluster answer}$
\EndIf
\State \Return answer
\end{algorithmic}
\end{algorithm}

\section{Experimental Setup}

\subsection{Models}

We evaluate 7 model pairs spanning 4 model families. Table~\ref{tab:models} summarizes the configurations. All experiments run on a single consumer machine (Apple Silicon) without quantization, demonstrating feasibility on commodity hardware.

\begin{table}[t]
\centering
\caption{Model pairs evaluated. ``Cross'' denotes cross-family routing where small and expert models are from different architectures.}
\label{tab:models}
\small
\begin{tabular}{@{}llcc@{}}
\toprule
\textbf{Small Model} & \textbf{Expert Model} & \textbf{Ratio} & \textbf{Type} \\
\midrule
Qwen2.5-0.5B & Qwen2.5-3B & 6$\times$ & Same \\
Qwen2.5-0.5B & Qwen2.5-1.5B & 3$\times$ & Same \\
Qwen2.5-0.5B & Phi-3.5-mini (3.8B) & 7.6$\times$ & Cross \\
Llama-3.2-1B & Llama-3.2-3B & 3$\times$ & Same \\
SmolLM2-360M & SmolLM2-1.7B & 4.7$\times$ & Same \\
SmolLM2-360M & Phi-3.5-mini (3.8B) & 10.6$\times$ & Cross \\
Gemma3-1B & Gemma3-4B & 4$\times$ & Same \\
\bottomrule
\end{tabular}
\end{table}

\subsection{Datasets}

We evaluate on 5 standard NLU benchmarks covering different reasoning capabilities:

\begin{itemize}[leftmargin=*]
    \item \textbf{BoolQ}~\cite{clark2019boolq}: Boolean question answering (yes/no)
    \item \textbf{HellaSwag}~\cite{zellers2019hellaswag}: Commonsense sentence completion (4-way MCQ)
    \item \textbf{ARC-Challenge}~\cite{clark2018arc}: Grade-school science reasoning (4-way MCQ)
    \item \textbf{ARC-Easy}~\cite{clark2018arc}: Easier subset of ARC (4-way MCQ)
    \item \textbf{WinoGrande}~\cite{sakaguchi2020winogrande}: Commonsense coreference resolution (2-way)
\end{itemize}

We use 50 examples per dataset for most model pairs, with 15 held-out examples for threshold calibration.

\subsection{Configuration}

\begin{figure}[t]
    \centering
    \includegraphics[width=\columnwidth]{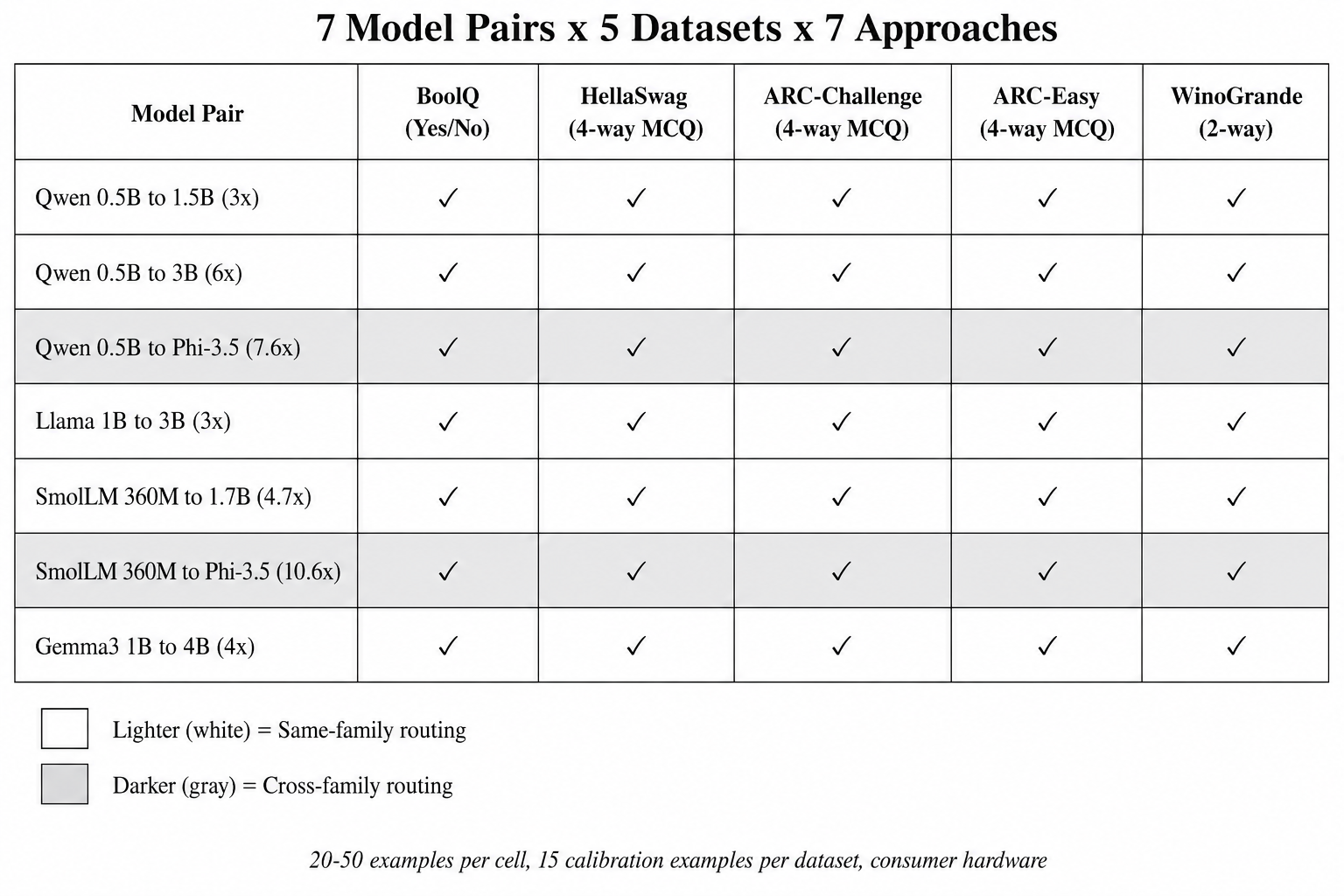}
    \caption{\textbf{Experimental coverage.} Each cell represents one dataset--model pair evaluated across all 7 approaches. Darker rows indicate cross-family configurations, where the expert is from a different architecture (Phi-3.5-mini). Sample sizes range from 20 to 50 per cell, with 15 held-out calibration examples per dataset.}
    \label{fig:exp_design}
\end{figure}

All experiments use temperature $T=0.7$, top-$p$ sampling with $p=0.95$, maximum generation length of 128 tokens, and top-$k=20$ logits for entropy computation. Semantic entropy approaches generate $N=5$ samples per query. Figure~\ref{fig:exp_design} summarizes the full experimental coverage.

\section{Results}

\subsection{Token Entropy is Blind in SLMs}

Our most striking finding is that token-level entropy provides no useful signal for small language models. Table~\ref{tab:token_entropy} shows the mean token entropy across all model pairs and datasets.

\begin{table}[t]
\centering
\caption{Mean token entropy (bits) for baseline generation. Values below 0.01 are effectively zero. 32 out of 35 combinations (91\%) show near-zero entropy.}
\label{tab:token_entropy}
\scriptsize
\setlength{\tabcolsep}{3pt}
\begin{tabular}{@{}lrrrrr@{}}
\toprule
\textbf{Pair} & \textbf{BQ} & \textbf{HS} & \textbf{AC} & \textbf{AE} & \textbf{WG} \\
\midrule
Qwen .5B to 3B & .000 & .000 & .001 & .000 & .000 \\
Qwen .5B to 1.5B & .000 & .000 & .000 & .000 & .000 \\
Qwen .5B to Phi & .000 & .000 & .000 & .000 & .000 \\
Llama 1B to 3B & .000 & .026 & .004 & .007 & .010 \\
SmolLM to 1.7B & .000 & .000 & .000 & .000 & .000 \\
SmolLM to Phi & .000 & .000 & .000 & .000 & .000 \\
Gemma3 1B to 4B & .000 & .000 & .000 & .000 & .000 \\
\bottomrule
\multicolumn{6}{@{}l}{\scriptsize BQ=BoolQ, HS=HellaSwag, AC=ARC-C,}\\
\multicolumn{6}{@{}l}{\scriptsize AE=ARC-E, WG=WinoGrande}\\
\end{tabular}
\end{table}

Figure~\ref{fig:token_entropy_blind} visualizes this collapse. SLMs produce extremely short, high-confidence outputs where every token has probability $>$0.999, yielding entropy indistinguishable from zero. Both correct and incorrect answers exhibit this same near-zero entropy, making threshold-based discrimination impossible: the calibrated threshold itself converges to zero.

This phenomenon arises because SLMs, when faced with questions they cannot reason through, tend to emit short, decisive answers (``A'', ``Yes'', ``No'') rather than extended reasoning chains. Each token in these short responses is produced with maximal confidence.

\subsection{Semantic Entropy Recovers a Useful Signal}

While SLMs are token-level confident, they are \emph{not} semantically consistent. When prompted multiple times with temperature sampling, an uncertain SLM produces different answers across runs. Table~\ref{tab:semantic_routing} presents the semantic routing results for all model pairs.

\begin{table*}[t]
\centering
\caption{Accuracy (\%) across all approaches for each model pair and dataset. Best result per row in \textbf{bold}. $\Delta$ shows semantic routing improvement over baseline.}
\label{tab:semantic_routing}
\scriptsize
\setlength{\tabcolsep}{4pt}
\begin{tabular}{@{}ll|ccccc|r@{}}
\toprule
\textbf{Pair} & \textbf{Dataset} & \textbf{Base} & \textbf{Tok-ES} & \textbf{Tok-Rt} & \textbf{Sem-ES} & \textbf{Sem-Rt} & \textbf{$\Delta$} \\
\midrule
\multirow{5}{*}{\shortstack[l]{Qwen 0.5B\\to 3B}}
 & BoolQ & 64 & \textbf{72} & 64 & 68 & 64 & 0 \\
 & HellaSwag & 24 & 42 & 28 & 32 & \textbf{62} & \color{gaingreen}{+38} \\
 & ARC-C & 36 & 32 & 38 & 34 & \textbf{64} & \color{gaingreen}{+28} \\
 & ARC-E & 20 & 20 & 22 & 18 & \textbf{48} & \color{gaingreen}{+28} \\
 & Wino. & 8 & 8 & 10 & 4 & \textbf{18} & \color{gaingreen}{+10} \\
\midrule
\multirow{5}{*}{\shortstack[l]{Qwen 0.5B\\to Phi-3.5}}
 & BoolQ & \textbf{70} & \textbf{70} & 65 & 60 & \textbf{70} & 0 \\
 & HellaSwag & 15 & 25 & 25 & \textbf{40} & 25 & +10 \\
 & ARC-C & 25 & 45 & 40 & 25 & \textbf{60} & \color{gaingreen}{+35} \\
 & ARC-E & 10 & 10 & 15 & 10 & \textbf{40} & \color{gaingreen}{+30} \\
 & Wino. & 5 & 5 & 5 & 5 & \textbf{20} & \color{gaingreen}{+15} \\
\midrule
\multirow{5}{*}{\shortstack[l]{SmolLM 360M\\to Phi-3.5}}
 & BoolQ & 80 & \textbf{85} & 75 & \textbf{85} & \textbf{85} & +5 \\
 & HellaSwag & 20 & 40 & 25 & 15 & \textbf{65} & \color{gaingreen}{+45} \\
 & ARC-C & 15 & 40 & 35 & 40 & \textbf{65} & \color{gaingreen}{+50} \\
 & ARC-E & 15 & 20 & 20 & 35 & \textbf{65} & \color{gaingreen}{+50} \\
 & Wino. & \textbf{70} & 55 & 60 & 55 & 50 & $-$20 \\
\midrule
\multirow{5}{*}{\shortstack[l]{SmolLM 360M\\to 1.7B}}
 & BoolQ & 80 & \textbf{85} & \textbf{85} & \textbf{85} & 75 & $-$5 \\
 & HellaSwag & 20 & \textbf{35} & 20 & 30 & 10 & $-$10 \\
 & ARC-C & 15 & 25 & 25 & \textbf{45} & 40 & +25 \\
 & ARC-E & 35 & 30 & 10 & 20 & \textbf{40} & +5 \\
 & Wino. & 40 & \textbf{65} & 45 & 45 & 40 & 0 \\
\midrule
\multirow{5}{*}{\shortstack[l]{Llama 1B\\to 3B}}
 & BoolQ & 60 & 45 & \textbf{70} & 55 & 60 & 0 \\
 & HellaSwag & 5 & 10 & \textbf{20} & 15 & \textbf{20} & +15 \\
 & ARC-C & 30 & 30 & \textbf{45} & 30 & 35 & +5 \\
 & ARC-E & 20 & 25 & \textbf{45} & 35 & 40 & +20 \\
 & Wino. & 0 & 0 & 0 & 0 & 0 & 0 \\
\midrule
\multirow{5}{*}{\shortstack[l]{Qwen 0.5B\\to 1.5B}}
 & BoolQ & \textbf{75} & 55 & 70 & 55 & 70 & $-$5 \\
 & HellaSwag & 35 & 35 & 35 & 20 & \textbf{50} & +15 \\
 & ARC-C & \textbf{45} & 25 & 30 & 35 & 40 & $-$5 \\
 & ARC-E & 10 & 5 & 10 & 10 & \textbf{30} & +20 \\
 & Wino. & 5 & 5 & 5 & 5 & 5 & 0 \\
\midrule
\multirow{5}{*}{\shortstack[l]{Gemma3 1B\\to 4B}}
 & BoolQ & \textbf{60} & \textbf{60} & \textbf{60} & \textbf{60} & \textbf{60} & 0 \\
 & ARC-E & \textbf{65} & \textbf{65} & \textbf{65} & 60 & \textbf{65} & 0 \\
 & Wino. & \textbf{55} & \textbf{55} & \textbf{55} & \textbf{55} & \textbf{55} & 0 \\
 & ARC-C & \textbf{45} & \textbf{45} & \textbf{45} & \textbf{45} & 40 & $-$5 \\
 & HellaSwag & \textbf{40} & 35 & 35 & 30 & 30 & $-$10 \\
\bottomrule
\end{tabular}
\end{table*}

Semantic routing wins the plurality of comparisons: it achieves the best accuracy in 13 out of 35 dataset--model pairs (37\%), followed by baseline at 11/35 (31\%), token early stop at 5/35 (14\%), token routing at 4/35 (11\%), and semantic early stop at 2/35 (6\%).

\begin{figure}[t]
    \centering
    \includegraphics[width=\columnwidth]{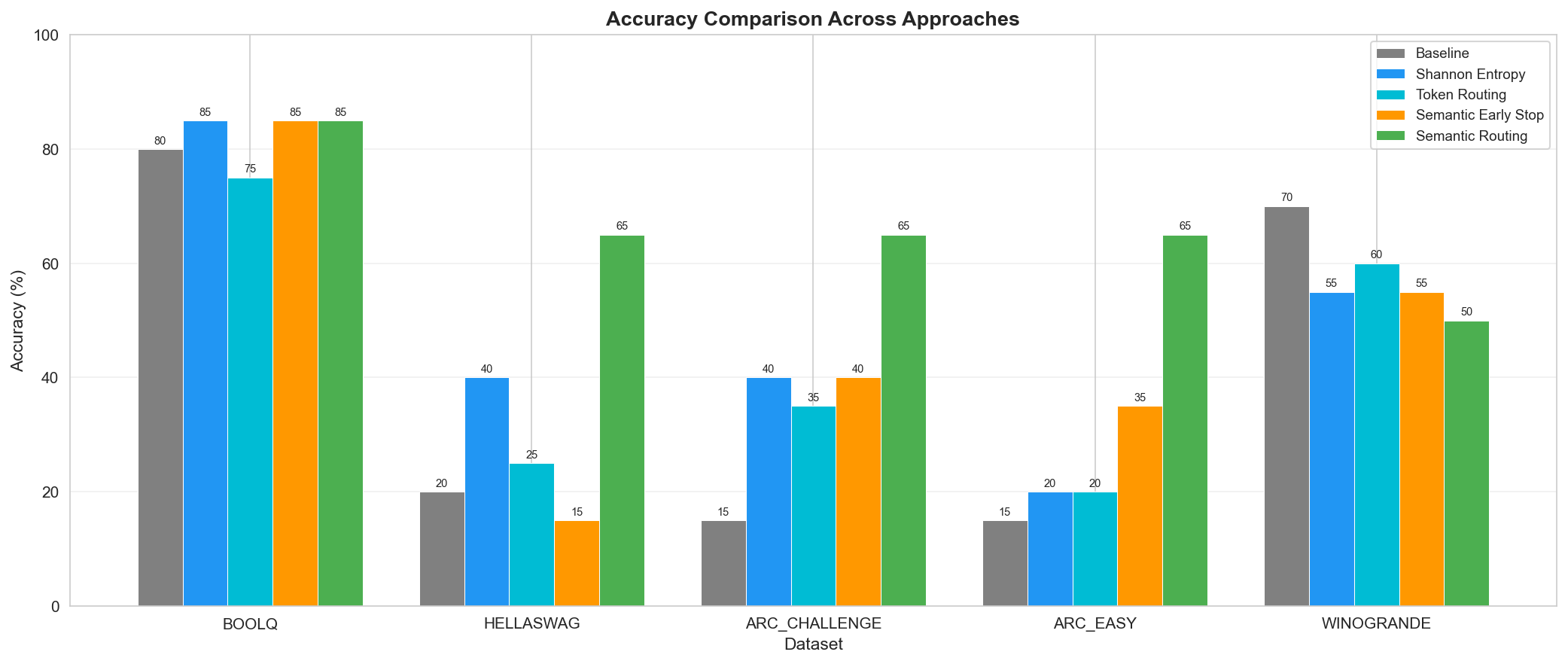}
    \caption{Accuracy across all approaches for SmolLM 360M to Phi-3.5-mini (cross-family routing to the best expert). Semantic Routing (rightmost bars) achieves the largest gains on reasoning-heavy datasets, with +50 percentage points over baseline on ARC-Challenge and ARC-Easy.}
    \label{fig:acc_smol}
\end{figure}

\begin{figure*}[t]
    \centering
    \includegraphics[width=0.85\textwidth]{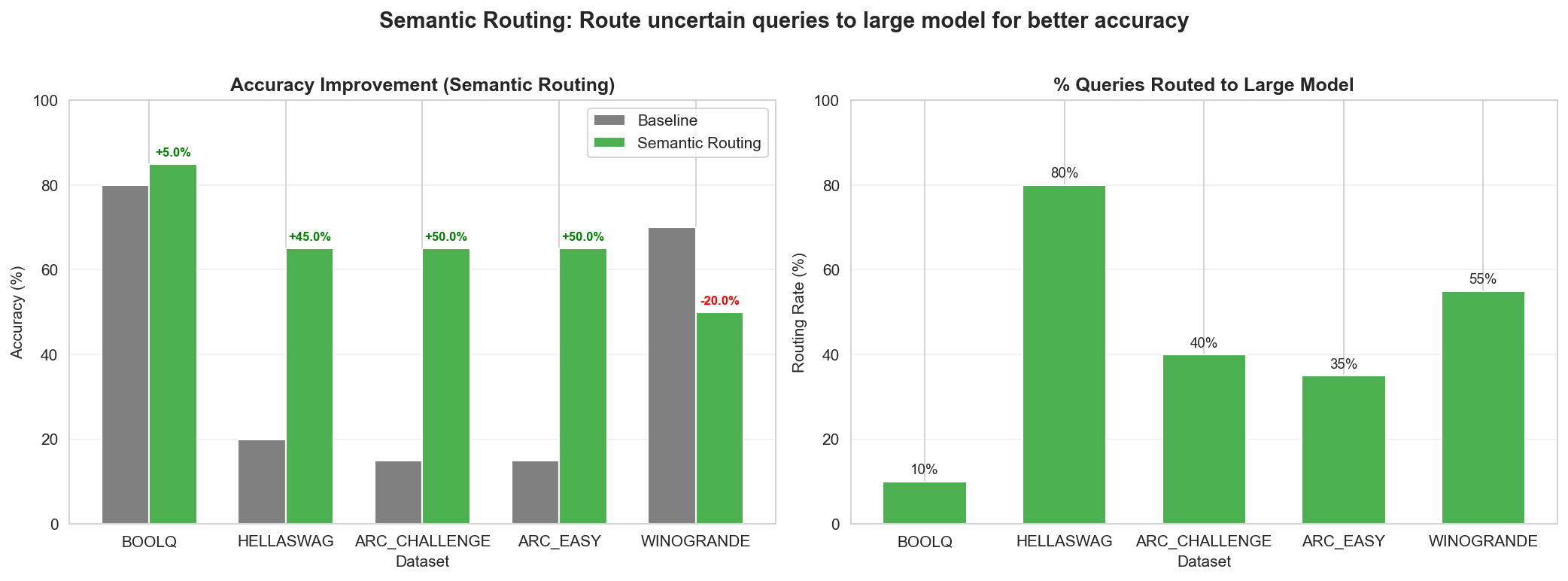}
    \caption{Semantic routing analysis for SmolLM 360M to Phi-3.5-mini. \textbf{Left:} Accuracy improvement over baseline, reaching +50\% on ARC-Challenge and ARC-Easy. \textbf{Right:} Routing rate, i.e., the fraction of queries where semantic entropy exceeded the threshold and the query was forwarded to the expert model. HellaSwag routes 80\% of queries, corresponding to the dataset where the small model is most uncertain.}
    \label{fig:semrt_detail}
\end{figure*}

Figure~\ref{fig:acc_smol} shows the accuracy comparison across all approaches for our strongest result (SmolLM~360M to Phi-3.5-mini), and Figure~\ref{fig:semrt_detail} shows the semantic routing mechanism in detail for this pair, illustrating both the accuracy improvement and the fraction of queries routed to the expert model per dataset.

\subsection{The Role of Expert Model Quality}

Perhaps our most practically significant finding is the dramatic difference between same-family and cross-family routing. Table~\ref{tab:routing_comparison} summarizes the average accuracy improvement from semantic routing across model pair types.

\begin{table}[t]
\centering
\caption{Average accuracy improvement ($\Delta$\%) from semantic routing over baseline, grouped by model pair.}
\label{tab:routing_comparison}
\small
\begin{tabular}{@{}lcc@{}}
\toprule
\textbf{Model Pair} & \textbf{Type} & \textbf{Avg $\Delta$} \\
\midrule
SmolLM 360M to Phi-3.5 & Cross & \textbf{+26.0\%} \\
Qwen 0.5B to Qwen 3B & Same & +20.8\% \\
Qwen 0.5B to Phi-3.5 & Cross & +18.0\% \\
Llama 1B to Llama 3B & Same & +8.0\% \\
Qwen 0.5B to Qwen 1.5B & Same & +5.0\% \\
SmolLM 360M to SmolLM 1.7B & Same & +3.0\% \\
Gemma3 1B to Gemma3 4B & Same & $-$3.0\% \\
\midrule
\textit{All cross-family} & Cross & \textbf{+22.0\%} \\
\textit{All same-family} & Same & +6.8\% \\
\bottomrule
\end{tabular}
\end{table}

\begin{figure}[t]
    \centering
    \includegraphics[width=\columnwidth]{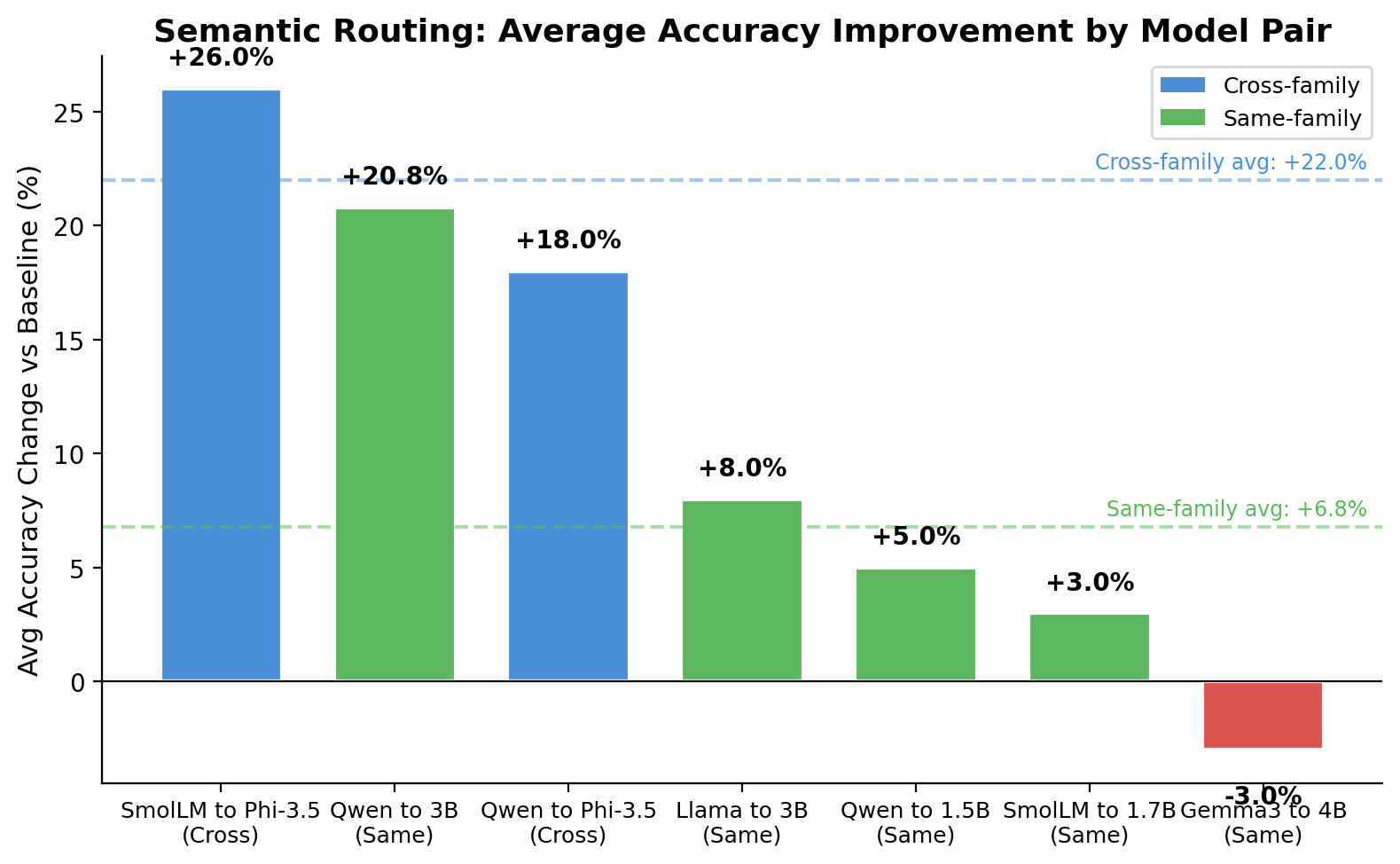}
    \caption{Average accuracy improvement from semantic routing across all 7 model pairs. Cross-family routing (blue) to a high-quality expert averages +22.0\% compared to +6.8\% for same-family routing (green). Dashed lines show group averages.}
    \label{fig:cross_vs_same}
\end{figure}

The contrast is striking: SmolLM~360M routed to its own larger sibling (1.7B) achieves only +3.0\% average improvement, but the \emph{same model} routed to Phi-3.5-mini achieves +26.0\%. The routing mechanism and uncertainty signal are identical; the only difference is the expert's quality. This suggests that deploying a single high-quality expert as a shared ``fallback'' for multiple weak SLMs is more effective than pairing each SLM with its family's larger variant. Figure~\ref{fig:cross_vs_same} visualizes this finding.

The Gemma3 results represent an instructive failure case: Gemma3-1B produces near-zero semantic entropy across all datasets (the model gives consistent but incorrect answers), so the routing signal never fires. This suggests that some model families may produce overconfident outputs even at the semantic level.

\subsection{Token Usage: Spending More, Not Less}

Semantic routing in SLMs \emph{always} uses more tokens than baseline. In all 35 dataset--model combinations, semantic routing consumed more compute, typically around 640 tokens per query compared to 2--120 tokens for baseline generation.

This increased cost comes from two sources: (1) generating $N=5$ samples for semantic entropy estimation, and (2) invoking the expert model for routed queries. In a typical configuration (Qwen~0.5B to 3B), baseline generation uses 30--120 tokens per query, while semantic routing consumes $\sim$640 tokens. The value proposition is not efficiency but accuracy: the additional compute is allocated precisely where the small model is uncertain, yielding targeted accuracy improvements.

\section{Analysis}

\subsection{Why Token Entropy Fails in SLMs}

The failure of token-level entropy in SLMs can be traced to a fundamental behavioral pattern. Small models tend to:

\begin{itemize}[leftmargin=*]
    \item Generate very short responses (often 2--5 tokens for MCQ tasks)
    \item Assign near-unity probability to each token in the sequence
    \item Skip reasoning entirely, jumping directly to an answer
\end{itemize}

As a result, even when the answer is incorrect, each token is produced with probability $>$0.999, yielding entropy $<$0.001 bits. The calibrated threshold converges to $\sim$0.0, eliminating any discriminative power.

\subsection{How Semantic Entropy Detects Uncertainty}

Semantic entropy circumvents this failure mode by probing the model's uncertainty at a different level. Consider a concrete example from our experiments (Qwen~0.5B on HellaSwag, correct answer: B):

\begin{itemize}[leftmargin=*]
    \item \textbf{Token entropy}: The model generates ``D'' with token entropy = 0.000 bits. The confidence signal suggests maximum certainty.
    \item \textbf{Semantic entropy}: Five samples yield answers \{D, A, B, D, C\}, producing four distinct clusters with probabilities $\{0.4, 0.2, 0.2, 0.2\}$, giving $H_{\text{semantic}} = 1.922$ bits. This exceeds the threshold (1.343 bits), triggering routing to Qwen~3B, which correctly answers ``B''.
\end{itemize}

The key insight is that while the model is confident about each \emph{token} it produces, repeated sampling reveals that it is \emph{not} confident about which \emph{answer} is correct. The model ``knows what it doesn't know'' at the answer level, even when it appears confident at the token level.

\subsection{Factors Affecting Routing Effectiveness}

We observe that routing effectiveness depends on several factors:

\textbf{Expert model capability.} The absolute quality of the expert model is the strongest predictor of routing success. Phi-3.5-mini, known for punching above its weight class, consistently yields the largest improvements as an expert.

\textbf{Small model uncertainty behavior.} Some models (e.g., Gemma3-1B) produce semantically consistent but incorrect answers, yielding low semantic entropy and preventing the routing signal from firing. Effective routing requires that the small model's uncertainty manifests as answer variation across samples.

\textbf{Dataset characteristics.} Routing benefits are largest for reasoning-heavy tasks (HellaSwag, ARC) where the expert model's additional capability translates to correctness. For simpler binary tasks (BoolQ) where the small model already performs well, routing provides minimal benefit.

\begin{figure*}[t]
    \centering
    \begin{subfigure}[t]{0.48\textwidth}
        \centering
        \includegraphics[width=\textwidth]{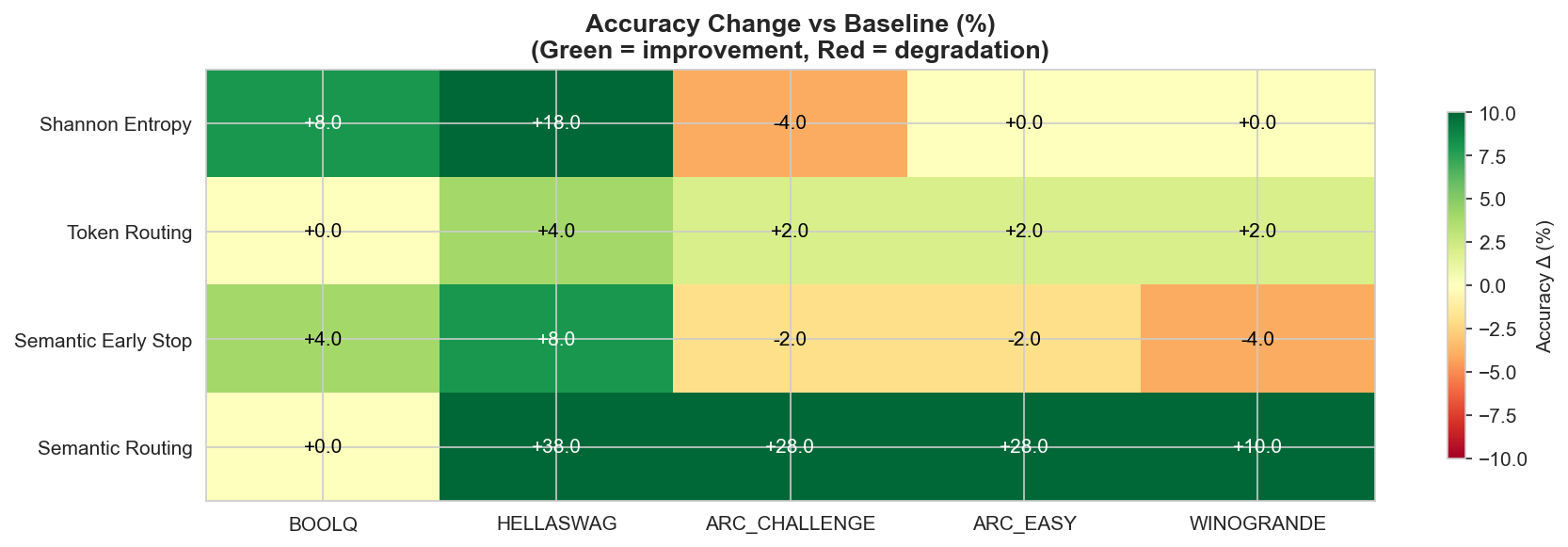}
        \caption{Qwen 0.5B to 3B: Semantic routing shows strong gains (dark green) across HellaSwag, ARC-Challenge, and ARC-Easy.}
        \label{fig:heatmap_qwen}
    \end{subfigure}
    \hfill
    \begin{subfigure}[t]{0.48\textwidth}
        \centering
        \includegraphics[width=\textwidth]{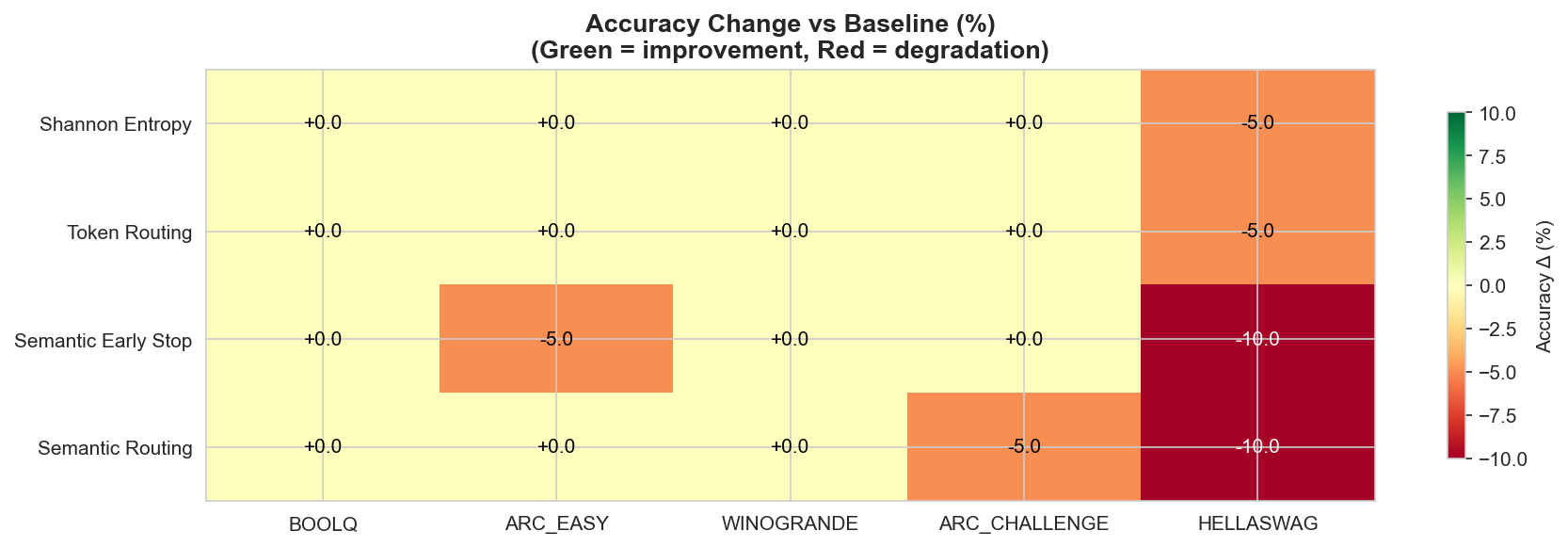}
        \caption{Gemma3 1B to 4B: No approach improves over baseline. Gemma3-1B produces overconfident outputs even at the semantic level.}
        \label{fig:heatmap_gemma}
    \end{subfigure}
    \caption{Accuracy change vs.\ baseline heatmaps comparing a successful routing pair (left) with a failure case (right). Green indicates improvement, orange/red indicates degradation.}
    \label{fig:heatmaps}
\end{figure*}

Figure~\ref{fig:heatmaps} contrasts the accuracy change heatmaps for a successful model pair (Qwen 0.5B to 3B) and a failure case (Gemma3 1B to 4B), highlighting how semantic routing effectiveness varies with model characteristics.

\subsection{Illustrative Examples}

Table~\ref{tab:examples} presents representative examples from our experiments showing the routing decision process in action.

\begin{table}[t]
\centering
\caption{Routing examples (Qwen 0.5B to 3B). Token entropy is 0.000 in all cases; semantic entropy correctly identifies uncertainty and triggers routing.}
\label{tab:examples}
\scriptsize
\setlength{\tabcolsep}{3pt}
\begin{tabular}{@{}lcccc@{}}
\toprule
\textbf{Dataset} & \textbf{Base} & \textbf{$H_{\text{sem}}$} & \textbf{$\tau$} & \textbf{Result} \\
\midrule
HellaSwag & \ding{55} (D) & 1.922 & 1.343 & \ding{51} Routed to B \\
ARC-C & \ding{55} & 0.722 & 0.361 & \ding{51} Routed to C \\
ARC-E & \ding{55} & 0.722 & 0.000 & \ding{51} Routed to B \\
BoolQ & \ding{55} (No) & 1.371 & 0.685 & \ding{51} Routed to Yes \\
\bottomrule
\end{tabular}

\vspace{6pt}
\raggedright
\textbf{Semantic entropy values} ($N\!=\!5$, exact-match clustering):\\[2pt]
\scriptsize
\begin{tabular}{@{}ll@{}}
5-0-0-0 (unanimous) = 0.000 & 2-2-1 (contested) = 1.522 \\
4-1 (one outlier) = 0.722 & 2-1-1-1 (scattered) = 1.922 \\
3-2 (majority split) = 0.971 & 1-1-1-1-1 (all differ) = 2.322 \\
3-1-1 (fragmented) = 1.371 & \\
\end{tabular}
\end{table}

\section{Discussion}

\subsection{Limitations}

\textbf{Sample size.} Our experiments use 20--50 examples per dataset--model combination. While sufficient for identifying trends and large effects, statistical significance of smaller differences ($<$10 percentage points) cannot be established at this scale. We present our findings as exploratory observations rather than statistically confirmed results.

\textbf{Compute overhead.} Semantic entropy requires $N=5$ forward passes through the small model plus (for routed queries) one pass through the expert model. For a 360M-parameter model routing to a 3.8B expert, this represents a $\sim$6$\times$ increase in total compute per query. The accuracy gains must be weighed against this cost in practical deployment.

\textbf{Model coverage.} Our study covers 4 model families (Qwen, Llama, SmolLM, Gemma) and 2 cross-family expert configurations. The Gemma3 failure case suggests that some architectures may not exhibit exploitable semantic uncertainty, and broader model coverage would help delineate when semantic routing is applicable.

\textbf{Clustering method.} Our bag-of-tokens embedding approach for free-text clustering is a simple baseline. More sophisticated methods (e.g., sentence-transformers) might yield better semantic grouping for non-MCQ tasks.

\subsection{Practical Implications}

Our findings suggest a practical deployment architecture for resource-constrained environments: run a small, fast model ($<$1B parameters) for all queries, and maintain a single high-quality expert (e.g., Phi-3.5-mini at 3.8B) as a shared fallback. Semantic entropy provides the routing signal, and the expert model need not be from the same family as the small model. This architecture enables:

\begin{itemize}[leftmargin=*]
    \item \textbf{Low latency for easy queries}: Most queries are answered by the fast small model without routing.
    \item \textbf{Accuracy recovery for hard queries}: Uncertain queries are identified via semantic entropy and routed to the expert.
    \item \textbf{Cost control}: The routing rate (what fraction of queries go to the expert) is controlled by the threshold $\tau$, allowing operators to trade accuracy for compute cost.
\end{itemize}

\subsection{Relationship to Prior Work}

Entropy-based confidence estimation has been studied in the context of larger models~\cite{sharma2025tjue, han2025overthinking}. Our work investigates whether these signals transfer to small models and finds that token-level entropy collapses, requiring a shift to semantic-level measurement. Semantic entropy~\cite{kuhn2023semantic} was originally proposed for uncertainty estimation; we demonstrate its utility as a \emph{routing signal} for small models, connecting it to model cascading~\cite{chen2023frugalgpt}. Our use of semantic entropy as the cascading criterion is, to our knowledge, new.

\section{Conclusion}

We investigated whether entropy-based confidence signals can help Small Language Models ($<$3B parameters) know what they don't know. Our exploratory study across 7 model pairs, 5 datasets, and 7 approaches reveals that token-level entropy is blind in SLMs, with 91\% of test configurations showing near-zero entropy regardless of correctness. Semantic entropy, which measures answer-level disagreement across multiple samples, recovers a viable uncertainty signal that enables effective routing to larger expert models, with accuracy improvements of up to +50 percentage points. The expert's quality matters more than family match: cross-family routing to a capable expert averages +22\% improvement versus +6.8\% for same-family routing.

The value proposition for entropy-based methods in SLMs is not about saving tokens; it is about spending them wisely. By identifying precisely when a small model is uncertain and selectively invoking a stronger model, we transform a limitation (the small model's ignorance) into an actionable signal for intelligent compute allocation.

\bibliographystyle{unsrt}

\end{document}